\documentclass[aos,MSNbibl,seceqn,dvips]{arximspdf}
\usepackage{stfloats}
\usepackage{graphicx}

\doi{10.1214/13-AOS1162} 
\volume{41}
\issue{6}
\pubyear{2013}
\firstpage{3022}
\lastpage{3049}

\makeatletter
  \fnbelowfloat

\newcommand{\rright}{\right}
\newcommand{\lleft}{\left}
\newcommand{\rrvert}{\vert}
\newcommand{\llvert}{\vert}
\newtheorem{theos}{Theorem}
\newproclaim{rem}{Remarks}
\newtheorem{lem}{Lemma}
\newproclaim{alg}{Algorithm}
\newtheorem{cor}{Corollary}
\newproclaim{exas}{Example}
\newtheorem{prop}{Proposition}
\newproclaim{defin}{Definition}
\newcommand{\eqref}[1]{(\ref{#1})}
\newcommand{\graph}{G}
\newcommand{\vertex}{V}
\newcommand{\edge}{E}
\newcommand{\defn}{:=}
\newcommand{\ko}{\perp\!\!\!\perp}
\newcommand{\compat}{\psi}
\newcommand{\statenum}{m}
\newcommand{\real}{\mathbb{R}}
\newcommand{\Exs}{\mathbb{E}}
\newcommand{\cov}{\operatorname{cov}}
\newcommand{\intr}{\operatorname{int}}
\newcommand{\trace}{\operatorname{trace}}
\newcommand{\graphtriang}{\widetilde{G}}
\newcommand{\Part}{\Phi}
\newcommand{\order}{\mathcal{O}}
\newcommand{\widebar}{\overline}
\makeatother

\begin{document}
\begin{frontmatter}

\title{Structure estimation for discrete graphical models: Generalized
covariance matrices and~their~inverses}
\runtitle{Structure estimation for discrete graphs}

\begin{aug}
\author{\fnms{Po-Ling} \snm{Loh}\corref{}\thanksref{t1,t2}\ead[label=e1]{ploh@berkeley.edu}}
\and
\author{\fnms{Martin J.} \snm{Wainwright}\thanksref{t2}\ead[label=e2]{wainwrig@stat.berkeley.edu}}
\runauthor{P.-L. Loh and M. J. Wainwright}
\affiliation{University of California, Berkeley}
\address{Department of Statistics\\
University of California, Berkeley\\
Berkeley, California 94720\\
USA \\
\printead{e1}\\
\phantom{E-mail: }\printead*{e2}} 
\end{aug}
\thankstext{t1}{Supported in part by a Hertz Foundation Fellowship and
a National Science Foundation Graduate Research Fellowship.}
\thankstext{t2}{Supported in part by NSF Grant DMS-09-07632 and Air
Force Office of Scientific Research Grant AFOSR-09NL184.}

\received{\smonth{2} \syear{2013}}
\revised{\smonth{7} \syear{2013}}

%
\begin{abstract}
We investigate the relationship between the structure of a discrete
graphical model and the support of the inverse of a generalized
covariance matrix. We show that for certain graph structures, the
support of the inverse covariance matrix of indicator variables on the
vertices of a graph reflects the conditional independence structure of
the graph. Our work extends results that have previously been
established only in the context of multivariate Gaussian graphical
models, thereby addressing an open question about the significance of
the inverse covariance matrix of a non-Gaussian distribution. The proof
exploits a combination of ideas from the geometry of exponential
families, junction tree theory and convex analysis. These
population-level results have various consequences for graph selection
methods, both known and novel, including a novel method for structure
estimation for missing or corrupted observations. We provide
nonasymptotic guarantees for such methods and illustrate the sharpness
of these predictions via simulations.
\end{abstract}

%
\pdfkeywords{62F12, 68W25, Graphical models, Markov random fields, model selection, inverse covariance estimation, high-dimensional statistics, exponential families, Legendre duality}

\end{frontmatter}

\section{Introduction}


Graphical models are used in many application domains, running the
gamut from computer vision and civil engineering to political science
and epidemiology. In many applications, estimating the edge structure
of an underlying graphical model is of significant interest. For
instance, a graphical model may be used to represent friendships
between people in a social network~\cite{BanEtal08} or links between
organisms with the propensity to spread an infectious
disease~\cite{NewWat99}. It is a classical corollary of the
Hammersley--Clifford theorem~\cite{Grimmett73,Besag74,Lau96} that zeros
in the inverse covariance matrix of a multivariate Gaussian
distribution indicate absent edges in the corresponding graphical
model. This fact, combined with various types of statistical estimators
suited to high dimensions, has been leveraged by many authors to
recover the structure of a Gaussian graphical model when the edge set
is sparse (see the papers~\cite{CaiEtal11,MeiBuh06,RavEtal11,Yua10} and
the references therein). Recently, Liu et
al.~\cite{LiuEtal12} and Liu, Lafferty and Wasserman \cite{LiuEtal09} introduced the notion of a~nonparanormal
distribution, which generalizes the Gaussian distribution by allowing
for monotonic univariate transformations, and argued that the same
structural properties of the inverse covariance matrix carry over to
the nonparanormal; see also the related work of Xue and
Zou~\cite{XueZou12} on copula transformations.

However, for non-Gaussian graphical models, the question of whether a
general relationship exists between conditional independence and the
structure of the inverse covariance matrix remains unresolved. In this
paper, we establish a number of interesting links between covariance
matrices and the edge structure of an underlying graph in the case of
discrete-valued random variables. (Although we specialize our treatment
to multinomial random variables due to their widespread applicability,
several of our results have straightforward generalizations to other
types of exponential families.) Instead of only analyzing the standard
covariance matrix, we show that it is often fruitful to augment the
usual covariance matrix with higher-order interaction terms. Our main
result has an interesting corollary for tree-structured graphs: for
such models, the inverse of a generalized covariance matrix is always
(block) graph-structured. In particular, for binary variables, the
inverse of the usual covariance matrix may be used to recover the edge
structure of the tree. We also establish more general results that
apply to arbitrary (nontree) graphs, specified in terms of graph
triangulations. This more general correspondence exploits ideas from
the geometry of exponential families~\cite{Brown86,WaiJor08}, as well
as the junction tree framework~\cite{Lauritzen88,Lau96}.

As we illustrate, these population-level results have a number of
corollaries for graph selection methods. Graph selection methods for
Gaussian data include neighborhood regression~\cite{MeiBuh06,Zhao06}
and the graphical Lasso~\cite{FriEtal08,RavEtal11,Rot08,AspBanGha08},
which corresponds to maximizing an $\ell_1$-regularized version of the
Gaussian likelihood. Alternative methods for selection of discrete
graphical models include the classical Chow--Liu algorithm for
trees~\cite{ChowLiu68}; techniques based on conditional entropy or
mutual information~\cite{AnaEtal11,BreEtal08}; and nodewise logistic
regression for discrete graphical models with pairwise
interactions~\cite{JalEtal11,RavEtal10}. Our population-level results
imply that minor variants of the graphical Lasso and neighborhood
regression methods, though originally developed for Gaussian data,
remain consistent for trees and the broader class of graphical models
with singleton separator sets. They also convey a cautionary message,
in that these methods will be inconsistent\vadjust{\goodbreak} (generically) for other
types of graphs. We also describe a new method for neighborhood
selection in an arbitrary sparse graph, based on linear regression over
subsets of variables. This method is most useful for bounded-degree
graphs with correlation decay, but less computationally tractable for
larger graphs.

In addition, we show that our methods for graph selection may be
adapted to handle noisy or missing data in a seamless manner. Naively
applying nodewise logistic regression when observations are
systematically corrupted yields estimates that are biased even in the
limit of infinite data. There are various corrections available, such
as multiple imputation~\cite{Rub87} and the expectation-maximization
(EM) algorithm~\cite{DemEtal77}, but, in general, these methods are not
guaranteed to be statistically consistent due to local optima. To the
best of our knowledge, our work provides the first method that is
provably consistent under high-dimensional scaling for estimating the
structure of discrete graphical models with corrupted observations.
Further background on corrupted data methods for low-dimensional
logistic regression may be found in Carroll, Ruppert and
Stefanski \cite{CarEtal95} and
Ibrahim et al.~\cite{IbrEtal05}.

The remainder of the paper is organized as follows. In
Section~\ref{SecBackground}, we provide brief background and notation on
graphical models and describe the classes of augmented covariance
matrices we will consider. In Section~\ref{SecResults}, we state our
main population-level result (Theorem~\ref{ThmFull}) on the
relationship between the support of generalized inverse covariance
matrices and the edge structure of a discrete graphical model, and then
develop a number of corollaries. The proof of Theorem~\ref{ThmFull} is
provided in Section~\ref{SecProofThmFull}, with proofs of corollaries
and more technical results deferred to the supplementary material \cite{LohWai13Sup}. In
Section~\ref{SecSample}, we develop consequences of our population-level
results in the context of specific methods for graphical model
selection. We provide simulation results in Section~\ref{SecSims} in
order to confirm the accuracy of our theoretically-predicted scaling
laws, dictating how many samples are required (as a function of graph
size and maximum degree) to recover the graph correctly.


\section{Background and problem setup}\label{SecBackground}
\setcounter{footnote}{2}

In this section, we provide background on graphical models and
exponential families. We then present a simple example illustrating
the phenomena and methodology underlying this paper.

\subsection{Undirected graphical models}

\begin{figure}
\begin{tabular}{@{}cc@{}}

\includegraphics{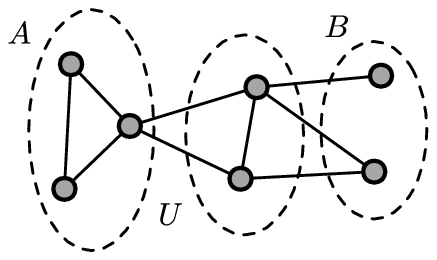}
 & \includegraphics{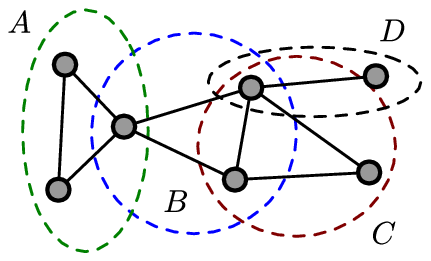}\\
\footnotesize{(a)} & \footnotesize{(b)}
\end{tabular}
\caption{\textup{(a)} Illustration of a vertex cutset: when the set
$U$ is removed, the graph breaks into two disjoint subsets
of vertices $A$ and $B$. \textup{(b)} Illustration of maximal cliques,
corresponding to fully connected subsets of vertices.}\label{FigGraphProps}
\end{figure}

An \emph{undirected graphical model} or \emph{Mar\-kov random field}
(MRF) is a family of probability distributions respecting the structure
of a fixed graph. We begin with some basic graph-theoretic terminology.
An undirected graph $\graph= (\vertex, \edge)$ consists of a collection
of vertices $\vertex= \{1, 2, \ldots, p\}$ and a collection of
unordered\footnote{No distinction is made between the edge $(s,t)$ and
the edge $(t,s)$. In this paper, we forbid graphs with self-loops,
meaning that $(s,s) \notin\edge$ for all $s \in\vertex$.} vertex pairs
$\edge\subseteq\vertex\times\vertex$. A~\emph{vertex cutset} is a
subset~$U$ of vertices whose removal breaks the graph into two or more
nonempty components [see Figure~\ref{FigGraphProps}(a)]. A
\emph{clique} is a subset $C \subseteq V$ such that $(s,t) \in\edge$
for all distinct $s,t \in C$. The cliques in
Figure~\ref{FigGraphProps}(b) are all \emph{maximal}, meaning they are
not properly contained within any other clique. For $s \in V$, we
define the neighborhood $N(s) \defn\{t \in\vertex\mid(s,t) \in\edge\}$
to be the set of vertices connected to $s$ by an edge.

For an undirected graph $G$, we associate to each vertex $s \in
\vertex$ a random variable~$X_s$ taking values in a space
$\mathcal{X}$. For any subset $A \subseteq\vertex$, we define $X_A
\defn\{X_s, s \in A \}$, and for three subsets of vertices, $A$, $B$
and $U$, we write $X_A \ko X_B \mid X_U$ to mean that the random vector
$X_A$ is conditionally independent of $X_B$ given $X_U$. The notion of
a Markov random field may be defined in terms of certain \emph{Markov
properties} indexed by vertex cutsets or in terms of a
\emph{factorization property} described by the graph cliques.

%

\begin{defin}[(Markov property)]
The random vector $X \defn(X_1, \ldots, X_p)$ is \emph{Markov with
respect to the graph $\graph$} if $X_A \ko X_B \mid X_U$ whenever $U$
is a vertex cutset that breaks the graph into disjoint subsets $A$ and
$B$.
\end{defin}

Note that the neighborhood set $N(s)$ is a vertex cutset for the sets
$A = \{s \}$ and $B = \vertex\setminus\{s \cup N(s)\}$. Consequently,
$X_s \ko X_{\vertex\setminus\{s \cup N(s)\}} \mid X_{N(s)}$. This
property is important for nodewise methods for graphical model
selection to be discussed later.

The factorization property is defined directly in terms of the
probability distribution $q$ of the random vector $X$. For each clique
$C$, a \emph{clique compatibility function} $\compat_C$ is a mapping
from configurations $x_C = \{x_s, s \in\vertex\}$ of variables to the
positive reals. Let $\mathcal{C}$ denote the set of all cliques in $G$.

\begin{defin}[(Factorization property)] The distribution of $X$ \emph{factorizes
according to $\graph$} if it may be written as a product of
clique functions:
%
\begin{equation}
\label{EqnClique} q(x_1, \ldots, x_p) \propto\prod
_{C \in
\mathcal{C}} \psi_C(x_C).
\end{equation}
\end{defin}

The factorization may always be restricted to maximal cliques of the
graph, but it is sometimes convenient to include terms for nonmaximal
cliques.

\subsection{Graphical models and exponential families}

By the Hammersley--Clifford theorem~\cite{Besag74,Grimmett73,Lau96},
the Markov and factorization properties are equivalent for any strictly
positive distribution. We focus on such strictly positive
distributions, in which case the factorization~\eqref{EqnClique} may
alternatively be represented in terms of an \emph{exponential family}
associated with the clique structure of $\graph$. We begin by defining
this exponential family representation for the special case of binary
variables ($\mathcal{X}= \{0,1\}$), before discussing a natural
generalization to $\statenum$-ary discrete random variables.

\subsubsection*{Binary variables} For a
binary random vector $X \in\{0,1\}^p$, we associate with each clique
$C$---both maximal and nonmaximal---a sufficient statistic
$\mathbb{I}_C(x_C) \defn\prod_{s \in C} x_s$. Note that $\mathbb
{I}_C(x_C) = 1$ if and only if $x_s = 1$ for all $s \in C$, so it is an
indicator function for the event $\{x_s = 1,\ \forall s \in C\}$. In
the exponential family, this sufficient statistic is weighted by a
natural parameter $\theta_C \in\real$, and we rewrite the
factorization~\eqref{EqnClique} as
%
\begin{eqnarray}
\label{EqnGeneralIsing} q_\theta(x_1, \ldots, x_p) & =&
\exp \biggl\{ \sum_{C
\in
\mathcal{C}} \theta_C
\mathbb{I}_C(x_C) - \Phi(\theta) \biggr\},
\end{eqnarray}
where $\Phi(\theta) \defn\log\sum_{x \in\{0,1\}^p} \exp( \sum_{C
\in\mathcal{C}} \theta_C \mathbb{I}_C(x_C))$ is the log normalization
constant. It may be verified (cf. Proposition 4.3 of Darroch and
Speed~\cite{DarSpe83}) that the factorization~\eqref{EqnGeneralIsing}
defines a minimal exponential family, that is, the statistics $\{
\mathbb{I}_C(x_C), C \in\mathcal{C}\}$ are affinely independent. In
the special case of pairwise interactions,
equation~\eqref{EqnGeneralIsing} reduces to the classical \emph{Ising
model}:
%
\begin{eqnarray}
\label{EqnIsing} q_\theta(x_1, \ldots, x_p) & =&
\exp \biggl\{ \sum_{s
\in
\vertex} \theta_s
x_s + \sum_{(s,t) \in\edge} \theta_{st}
x_s x_t - \Phi(\theta) \biggr\}.
\end{eqnarray}
The model~\eqref{EqnIsing} is a particular instance of a pairwise
Markov random field.

\subsubsection*{Multinomial variables} In order
to generalize the Ising model to nonbinary variables, say, $\mathcal{X}
= \{0,1, \ldots, \statenum-1\}$, we introduce a larger set of
sufficient statistics. We first illustrate this extension for a
pairwise Markov random field. For each node $s \in\vertex$ and
configuration $j \in\mathcal{X}_0 \defn\mathcal{X} \setminus\{0\} =
\{1, 2, \ldots, \statenum-1\}$, we introduce the binary-valued
indicator function
%
\begin{equation}
\label{EqnBinaryIndicator} \mathbb{I}_{s;j}(x_s)= \cases{1, &\quad if
$x_s = j$,
\cr
0, &\quad otherwise.}
\end{equation}
We also introduce a vector $\theta_s = \{ \theta_{s;j}, j \in
\mathcal{X}_0 \}$ of natural parameters associated with these
sufficient statistics. Similarly, for each edge $(s,t) \in\edge$ and
configuration $(j,k) \in\mathcal{X}^2_0 \defn\mathcal{X}_0 \times
\mathcal{X}_0$, we introduce the binary-valued indicator function
$\mathbb{I}_{st; jk}$ for the event $\{x_s = j, x_t =k \}$, as well as
the collection $\theta_{st} \defn\{ \theta_{st; jk}, (j,k) \in
\mathcal{X}^2_0 \}$ of natural parameters. Then any pairwise Markov
random field over $\statenum$-ary random variables may be written in
the form
%
\begin{eqnarray}
\label{EqnPairwiseDiscrete}\qquad q_\theta(x_1, \ldots, x_p) & =&
\exp \biggl\{ \sum_{s
\in\vertex} \bigl\langle
\theta_s, \mathbb{I}_s(x_s) \bigr\rangle+
\sum_{(s,t) \in
\edge} \bigl\langle\theta_{st},
\mathbb{I}_{st}(x_s, x_t) \bigr\rangle- \Phi(
\theta) \biggr\},
\end{eqnarray}
where\vspace*{-1pt} we have used the shorthand $\langle\theta_s,
\mathbb{I}_s(x_s)\rangle
\defn\sum_{j=1}^{\statenum-1} \theta_{s;j} \mathbb{I}_{s;j}(x_s)$ and
$\langle\theta_{st},\break  \mathbb{I}_{st}(x_s, x_t)\rangle
\defn\sum_{j,k=1}^{\statenum-1} \theta_{st; jk} \mathbb{I}_{st;
jk}(x_s, x_t)$. Equation~\eqref{EqnPairwiseDiscrete} defines a minimal
exponential family with dimension $|\vertex| (\statenum-1) + |\edge|
(\statenum-1)^2$~\cite{DarSpe83}. Note that the
family~\eqref{EqnPairwiseDiscrete} is a natural generalization of the
Ising model~\eqref{EqnIsing}; in particular, when $\statenum=2$, we
have a single sufficient statistic $\mathbb{I}_{s;1}(x_s) = x_s$ for
each vertex and a~single sufficient statistic $\mathbb{I}_{st;11}(x_s,
x_t) = x_s x_t$ for each edge. (We have omitted the additional
subscript $1$ or $11$ in our earlier notation for the Ising model,
since they are superfluous in that case.)

Finally, for a graphical model involving higher-order interactions, we
require additional sufficient statistics. For each clique $C \in
\mathcal{C}$, we define the subset of configurations
\[
\mathcal{X}^{|C|}_0 \defn\underbrace{
\mathcal{X}_0 \times\cdots\times\mathcal{X}_0}_{C\ \mathrm{times}}
= \bigl\{ (j_s, s \in C) \in\mathcal{X}^{|C|} \dvtx
j_s \neq0\ \forall s \in C \bigr\},
\]
a set of cardinality $(\statenum-1)^{|C|}$. As before, $\mathcal{C}$ is
the set of all maximal and nonmaximal cliques. For any configuration
$J = \{j_s, s \in C\} \in\mathcal{X}_0^{|C|}$, we define the corresponding
indicator function
%
\begin{equation}
\label{EqnSuff} \mathbb{I}_{C;J}(x_C) = \cases{ 1, &\quad if
$x_C = J$,
\cr
0, &\quad otherwise.}
\end{equation}
We then consider the general multinomial exponential family
%
\begin{eqnarray}\label{EqnGeneralMultinomial}
q_\theta(x_1, \ldots, x_p)  &=& \exp \biggl\{ \sum_{C
\in
\mathcal{C}} \langle\theta_C,
\mathbb{I}_C\rangle- \Phi(\theta) \biggr\}
\nonumber\\[-12pt]\\[-8pt]
\eqntext{\mbox{for } x_s \in\mathcal{X}= \{0,1, \ldots, \statenum-1\}}
\end{eqnarray}
with $\langle\theta_C, \mathbb{I}_C(x_C)\rangle = \sum_{J \in\mathcal
{X}_0^{|C|}} \theta_{C;J} \mathbb{I}_{C;J}(x_C)$. Note that our
previous models---namely, the binary models~\eqref{EqnGeneralIsing}
and~\eqref{EqnIsing}, as well as the pairwise multinomial
model~\eqref{EqnPairwiseDiscrete}---are special cases of this general
factorization.

Recall that an exponential family is \emph{minimal} if no nontrivial
linear combination of sufficient statistics is almost surely equal to a
constant. The family is \emph{regular} if $\{\theta\dvtx  \Phi(\theta)
< \infty\}$ is an open set. As will be relevant later, the exponential
families described in this section are all minimal and
regular~\cite{DarSpe83}.
%

\subsection{Covariance matrices and beyond}
\label{SecExamples}

We now turn to a discussion of the phenomena that motivate the analysis
of this paper. Consider the usual covariance matrix $\Sigma= \cov(X_1,
\ldots, X_p)$. When $X$ is jointly Gaussian, it is an immediate
consequence of the Hammersley--Clifford theorem that the sparsity
pattern of the precision matrix $\Gamma= \Sigma^{-1}$ reflects the
graph structure---that is, $\Gamma_{st} = 0$ whenever $(s,t)
\notin\edge$. More precisely, $\Gamma_{st}$ is a scalar multiple of the
correlation of $X_s$ and $X_t$ conditioned on $X_{\setminus\{s,t\}}$
(cf. Lauritzen~\cite{Lau96}). For non-Gaussian distributions, however,
the conditional correlation will be a function of
$X_{\setminus\{s,t\}}$, and it is unknown whether the entries of
$\Gamma$ have any relationship with the strengths of correlations along
edges in the graph.

Nonetheless, it is tempting to conjecture that inverse covariance
matrices might be related to graph structure
in the non-Gaussian case. We explore this possibility by
considering a simple case of the binary Ising model~\eqref{EqnIsing}.

\begin{exas}
\label{ExasGraph} Consider a simple chain graph on four nodes, as
illustrated in Figure~\ref{FigGraphs}(a). In terms of the
factorization~\eqref{EqnIsing}, let the node potentials be $\theta_s =
0.1$ for all $s \in V$ and the edge potentials be $\theta_{st} = 2$ for
all $(s,t) \in E$. For a multivariate Gaussian graphical model defined
on $G$, standard theory predicts that the inverse covariance matrix
$\Gamma= \Sigma^{-1}$ of the distribution is graph-structured:
$\Gamma_{st} = 0$ if and only if $(s,t) \notin E$. Surprisingly, this
is also the case for the chain graph with binary variables [see
panel~(f)]. However, this statement is \emph{not} true for the
single-cycle graph shown in panel~(b). Indeed, as shown in panel~(g),
the inverse covariance matrix has no nonzero entries at all. Curiously,
for the more complicated graph in~(e), we again observe a
graph-structured inverse covariance matrix.

%
\begin{figure}
\begin{tabular}{@{}ccccc@{}}

\includegraphics{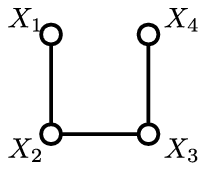}
 & \includegraphics{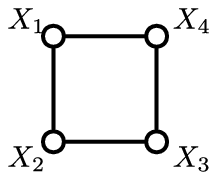} & \includegraphics{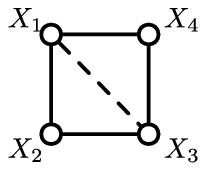} & \includegraphics{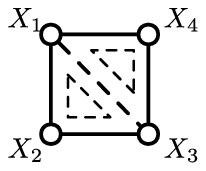} & \includegraphics{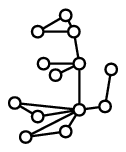}\\
\footnotesize{(a) Chain} & \footnotesize{(b) Single cycle} & \footnotesize{(c) Edge augmented} & \footnotesize{(d) With 3-cliques} & \footnotesize{(e) Dino}
\end{tabular}\vspace*{12pt}
\begin{tabular}{cc}
\footnotesize{$\Gamma_{\mathrm{chain}} =
\left[\matrix{
9.80 & -3.59 & 0 & 0 \cr
-3.59 & 34.30 & -4.77 & 0 \cr
0 & -4.77 & 34.30 & -3.59 \cr
0 & 0 & -3.59 & 9.80}\right]$}
& 
\footnotesize{$\Gamma_{\mathrm{loop}} =
\left[\matrix{
51.37 & -5.37 & -0.17 & -5.37 \cr
-5.37 & 51.37 & -5.37 & -0.17 \cr
-0.17 & -5.37 & 51.37 & -5.37 \cr
-5.37 & -0.17 & -5.37 & 51.37}\right]$}
\\[12pt]
\footnotesize{(f)} & \footnotesize{(g)}
\end{tabular}
\caption{\textup{(a)--(e)} Different examples of graphical models. \textup{(f)} Inverse
covariance for chain graph in~\textup{(a)}. \textup{(g)} Inverse covariance for
single-cycle graph in~\textup{(b)}.}\label{FigGraphs}
\end{figure}

Still focusing on the single-cycle graph in panel~(b), suppose that
instead of considering the ordinary covariance matrix, we compute the
covariance matrix of the \emph{augmented} random vector $(X_1, X_2,
X_3, X_4, X_1 X_3)$, where the extra term $X_1 X_3$ is represented by
the dotted\vadjust{\goodbreak} edge shown in panel~(c). The $5 \times5$ inverse of this
generalized covariance matrix takes the form
%
\begin{equation}
\label{EqnAugmented} \Gamma_{\mathrm{aug}} = 10^3 \times\lleft[
\matrix{1.15 & -0.02 & 1.09 & -0.02 & -1.14
\cr
-0.02 & 0.05 & -0.02 & 0 & 0.01
\cr
1.09 & -0.02 & 1.14 & -0.02 & -1.14
\cr
-0.02 & 0 & -0.02 & 0.05 & 0.01
\cr
-1.14 & 0.01 & -1.14 & 0.01 & 1.19} \rright].
\end{equation}
This matrix safely separates nodes $1$ and $4$, but the entry
corresponding to the nonedge $(1,3)$ is \emph{not} equal to
zero. Indeed, we would observe a similar phenomenon if we chose to
augment the graph by including the edge $(2,4)$ rather than
$(1,3)$.
This example shows that the usual inverse covariance
matrix is not always graph-structured, but inverses of augmented
matrices involving higher-order interaction terms may
reveal graph structure.

Now let us consider a more general graphical model that adds the
3-clique interaction terms shown in panel~(d) to the usual Ising terms.
We compute the covariance matrix of the augmented vector
\begin{eqnarray*}
\Psi(X) & =& \{X_1, X_2, X_3,
X_4, X_1 X_2, X_2
X_3, X_3 X_4,
\\
&&\hspace*{23pt} X_1 X_4, X_1 X_3,
X_1 X_2 X_3, X_1
X_3 X_4 \} \in\{0,1 \}^{11}.
\end{eqnarray*}
Empirically, one may show that the $11 \times11$ inverse
$(\cov[\Psi(X)])^{-1}$ respects aspects of the graph structure: there
are zeros in position $(\alpha, \beta)$, corresponding to the
associated functions $X_\alpha= \prod_{s \in\alpha} X_s$ and $X_\beta=
\prod_{s \in\beta} X_\beta$, whenever $\alpha$ and $\beta$ do not lie
within the same maximal clique. [E.g., this applies to the
pairs $(\alpha, \beta) = (\{2\}, \{4\})$ and $(\alpha, \beta) = (\{2\},
\{1,4\})$.]
\end{exas}

The goal of this paper is to understand when certain inverse
covariances do (and \emph{do not}) capture the structure of a graphical
model. At its root is the principle that the augmented inverse
covariance matrix $\Gamma= \Sigma^{-1}$, suitably defined, is
\emph{always} graph-structured with respect to a graph triangulation.
In some cases [e.g., the dino graph in Figure~\ref{FigGraphs}(e)], we
may leverage the block-matrix inversion formula~\cite{HorJoh90},
namely,
%
\begin{equation}
\label{EqnChai} \Sigma_{A,A}^{-1} = \Gamma_{A,A} -
\Gamma_{A,B} \Gamma_{B,B}^{-1} \Gamma_{B,A},
\end{equation}
to conclude that the inverse of a sub-block of the augmented matrix
(e.g., the ordinary covariance matrix) is still graph-structured. This
relation holds whenever $A$~and~$B$ are chosen in such a way that the
second term in equation~\eqref{EqnChai} continues to respect the edge
structure of the graph. These ideas will be made rigorous in
Theorem~\ref{ThmFull} and its corollaries in the next section.

\section{Generalized covariance matrices and graph structure}
\label{SecResults}

We now state our main results on the relationship between the zero
pattern of generalized (augmented) inverse covariance matrices and graph
structure. In Section~\ref{SecSample} to follow, we develop some
consequences of these results for data-dependent estimators
used in structure estimation.

We begin with some notation for defining generalized covariance
matrices, stated in terms of the sufficient statistics previously
defined~\eqref{EqnSuff}. Recall that a~clique $C \in \mathcal{C}$ is
associated with the collection $\{\mathbb{I}_{C;J}, J
\in\mathcal{X}_0^{|C|}\}$ of binary-valued sufficient statistics. Let
$\mathcal{S} \subseteq\mathcal{C}$, and define the random vector
%
\begin{eqnarray}
\Psi(X; \mathcal{S}) & =& \bigl\{ \mathbb{I}_{C;J}, J \in\mathcal
{X}_0^{|C|}, C \in\mathcal{S} \bigr\},
\end{eqnarray}
consisting of all the sufficient statistics indexed by elements of
$\mathcal{S}$. As in the previous section, the set $\mathcal{C}$ contains both
maximal and nonmaximal cliques.

We will often be interested in situations where $\mathcal{S}$ contains
all subsets of a given set. For a subset $A \subseteq\vertex$, let
$\operatorname{pow}(A)$ denote the collection of all $2^{|A|} - 1$
nonempty subsets of $A$. We extend this notation to $\mathcal{S}$ by
defining
\[
\operatorname{pow}(\mathcal{S}) \defn\bigcup_{C \in\mathcal{S}}
\operatorname{pow}(C).
\]

\subsection{Triangulation and block structure}

Our first main result concerns a connection between the inverses of
generalized inverse covariance matrices associated with the
model~\eqref{EqnGeneralMultinomial} and any triangulation of the
underlying graph $\graph$. The notion of a triangulation is defined in
terms of chordless cycles, which are sequences of distinct vertices
$\{s_1, \ldots, s_\ell\}$ such that:
\begin{itemize}
\item $(s_i, s_{i+1}) \in\edge$ for all $1 \le i \le\ell- 1$, and
    also $(s_\ell, s_1) \in\edge$;

\item no other nodes in the cycle are connected by an edge.
\end{itemize}
As an illustration, the $4$-cycle in Figure~\ref{FigGraphs}(b) is a
chordless cycle.

\begin{defin}[(Triangulation)]
Given an undirected graph $\graph= (\vertex, \edge)$, a
\emph{triangulation} is an augmented graph $\graphtriang= (\vertex,
\widetilde{\edge}) $ that contains no chordless cycles of length
greater than 3.
\end{defin}

Note that a tree is trivially triangulated, since it contains no
cycles. On the other hand, the chordless $4$-cycle in
Figure~\ref{FigGraphs}(b) is the simplest example of a nontriangulated
graph. By adding the single edge $(1,3)$ to form the augmented edge set
$\widetilde{\edge}= \edge\cup\{(1,3)\}$, we obtain the triangulated
graph $\graphtriang= (\vertex, \widetilde{\edge})$ shown in panel~(c).
One may check that the more complicated graph shown in
Figure~\ref{FigGraphs}(e) is triangulated as well.

Our first result concerns the inverse $\Gamma$ of the matrix
$\cov(\Psi(X; \widetilde{\mathcal{C}}))$, where $\widetilde
{\mathcal{C}}$ is the set of all cliques arising from some
triangulation $\widetilde{G}$ of $\graph$. For any two subsets $A, B
\in\widetilde{\mathcal{C}}$, we write $\Gamma(A,B)$ to denote the
sub-block of $\Gamma$ indexed by all indicator statistics on $A$ and
$B$, respectively. (Note that we are working with respect to the
exponential family representation over the triangulated graph
$\widetilde{G}$.) Given our previously-defined sufficient\vadjust{\goodbreak}
statistics~\eqref{EqnSuff}, the sub-block $\Gamma(A,B)$ has dimensions
$d_A \times d_B$, where
\[
d_A \defn(m-1)^{|A|}\quad\mbox{and}\quad d_B
\defn(m-1)^{|B|}.
\]
For example, when $A = \{s\}$ and $B = \{t\}$, the
submatrix $\Gamma(A,B)$ has dimension $(\statenum-1) \times
(\statenum-1)$. With this notation, we have the following result:

\begin{theos}[(Triangulation and block graph-structure)]\label{ThmFull}
Consider an arbitrary
discrete graphical model of the form~\eqref{EqnGeneralMultinomial}, and
let $\widetilde{\mathcal {C}}$ be the set of all cliques in any
triangulation of $\graph$. Then the generalized covariance matrix
$\cov(\Psi(X; \widetilde{\mathcal{C}}))$ is invertible, and its inverse
$\Gamma$ is \emph{block graph-structured}:
\begin{enumerate}[(a)]
\item[(a)] For any two subsets $A, B \in\widetilde{\mathcal{C}}$
    that are not subsets of the same maximal clique, the block
    $\Gamma(A, B)$ is identically zero.

\item[(b)] For almost all parameters $\theta$, the entire block
$\Gamma(A,B)$ is nonzero whenever $A$~and~$B$ belong to a common
maximal clique.
\end{enumerate}
\end{theos}

In part (b), ``almost all'' refers to all parameters $\theta$ apart
from a set of Lebesgue measure zero. The proof of
Theorem~\ref{ThmFull}, which we provide in
Section~\ref{SecProofThmFull}, relies on the geometry of exponential
families~\cite{Brown86,WaiJor08} and certain aspects of convex
analysis~\cite{Roc70}, involving the log partition function $\Phi$ and
its Fenchel--Legendre dual $\Phi^*$. Although we have stated
Theorem~\ref{ThmFull} for discrete variables, it easily generalizes to
other classes of random variables. The only difference is the specific
choices of sufficient statistics used to define the generalized
covariance matrix. This generality becomes apparent in the proof.

To provide intuition for Theorem~\ref{ThmFull}, we consider its
consequences for specific graphs. When the original graph is a tree
[such as the graph in Figure~\ref{FigGraphs}(a)], it is already
triangulated, so the set $\widetilde{\mathcal{C}}$ is equal to the edge
set $E$, together with singleton nodes. Hence, Theorem~\ref{ThmFull}
implies that the inverse $\Gamma$ of the matrix of sufficient
statistics for vertices and edges is graph-structured, and blocks of
nonzeros in $\Gamma$ correspond to edges in the graph. In particular,
we may apply Theorem~\ref{ThmFull}(a) to the subsets $A = \{s \}$ and
$B = \{t\}$, where $s$ and $t$ are distinct vertices with $(s,t)
\notin\edge$, and conclude that the $(\statenum-1) \times
(\statenum-1)$ sub-block $\Gamma(A,B)$ is equal to zero.

When $G$ is not triangulated, however, we may need to invert a larger
augmented covariance matrix and include sufficient statistics over
pairs $(s,t) \notin E$ as well. For instance, the augmented graph shown
in Figure~\ref{FigGraphs}(c) is a triangulation of the chordless
$4$-cycle in panel~(b). The associated set of maximal cliques is given
by $\widebar{\mathcal{C}}= \{(1,2), (2,3), (3,4), (1,4), (1,3) \}$;
among other predictions, our theory guarantees that the generalized
inverse covariance $\Gamma$ will have zeros in the sub-block
$\Gamma(\{2\}, \{4\})$.

\subsection{Separator sets and graph structure}

In fact, it is not necessary to take sufficient statistics
over \emph{all} maximal cliques, and we may consider a slightly smaller
augmented covariance matrix. (This simpler type of augmented
covariance matrix explains the calculations given in
Section~\ref{SecExamples}.)\vadjust{\goodbreak}

By classical graph theory, any triangulation $\widetilde{G}$ gives rise
to a \emph{junction tree} representation of $\graph$. Nodes in the
junction tree are subsets of $\vertex$ corresponding to maximal cliques
of $\widetilde{G}$, and the intersection of any two adjacent cliques
$C_1$ and $C_2$ is referred to as a \emph{separator set} $S = C_1 \cap
C_2$. Furthermore, any junction tree must satisfy the \emph{running
intersection property}, meaning that for any two nodes of the junction
tree---say, corresponding to cliques $C$ and $D$---the intersection $C
\cap D$ must belong to every separator set on the unique path between
$C$ and $D$. The following result shows that it suffices to construct
generalized covariance matrices augmented by separator sets:

\begin{cor}
\label{CorSep} Let $\mathcal{S}$ be the set of separator sets in any
triangulation of $\graph$, and let $\Gamma$ be the inverse of
$\cov(\Psi(X; \vertex\cup \operatorname{pow}(\mathcal{S})))$. Then
$\Gamma(\{s\}, \{t\}) = 0$ whenever $(s,t) \notin\widetilde{\edge}$.
\end{cor}

Note that $V \cup\operatorname{pow}(\mathcal{S}) \subseteq\widetilde
{\mathcal {C}}$, and the set of sufficient statistics considered in
Corollary~\ref{CorSep} is generally much smaller than the set of
sufficient statistics considered in Theorem~\ref{ThmFull}. Hence, the
generalized covariance matrix of Corollary~\ref{CorSep} has a smaller
dimension than the generalized covariance matrix of
Theorem~\ref{ThmFull}, which becomes significant when we consider
exploiting these population-level results for statistical estimation.

The graph in Figure~\ref{FigGraphs}(c) of Example~\ref{ExasGraph} and
the associated matrix in equation~\eqref{EqnAugmented} provide a
concrete example of Corollary~\ref{CorSep} in action. In this case, the
single separator set in the triangulation is $\{1, 3\}$, so when
$\mathcal{X}= \{0, 1\}$, augmenting the usual covariance matrix with
the additional sufficient statistic $\mathbb{I}_{13; 11}(x_1, x_3) =
x_1 x_3$ and taking the inverse yields a graph-structured matrix.
Indeed, since $(2,4) \notin\widetilde{E}$, we observe that
$\Gamma_{\mathrm{aug}}(2,4) = 0$ in equation~\eqref{EqnAugmented},
consistent with the result of Corollary~\ref{CorSep}.

Although Theorem~\ref{ThmFull} and Corollary~\ref{CorSep} are clean
population-level results, however, forming an appropriate augmented
covariance matrix requires prior knowledge of the graph, namely, which
edges are involved in a suitable triangulation. This is infeasible in
settings where the goal is to recover the edge structure of the graph.
Corollary~\ref{CorSep} is most useful for edge recovery when $G$ admits
a triangulation with only singleton separator sets, since then $V
\cup\operatorname{pow}(\mathcal{S}) = V$. In particular, this condition
holds when $G$ is a tree. The following corollary summarizes our
result:
%
\begin{cor}
\label{CorTree}
For any graph with singleton separator sets, the inverse $\Gamma$ of
the covariance matrix $\cov(\Psi(X; \vertex))$ of vertex statistics is
graph-structured. (This class includes trees as a special case.)
\end{cor}

In the special case of binary variables, we have $\Psi(X; \vertex) =
(X_1, \ldots, X_p)$, so Corollary~\ref{CorTree} implies that the
inverse of the ordinary covariance matrix $\cov(X)$ is
graph-structured. For $m$-ary variables, $\cov(\Psi(X;V))$ is a matrix
of dimensions $(m-1)p \times(m-1)p$ involving indicator\vadjust{\goodbreak} functions for
each variable. Again, we may relate this corollary to
Example~\ref{ExasGraph}---the inverse covariance matrices for the tree
graph in panel~(a) and the dino graph in panel~(e) are exactly
graph-structured. Although the dino graph is not a tree, it possesses
the nice property that the only separator sets in its junction tree are
singletons.

Corollary~\ref{CorSep} also guarantees that inverse covariances may be
partially graph-structured, in the sense that $\Gamma(\{s\}, \{t\}) =
0$ for any pair of vertices $(s,t)$ separable by a singleton separator
set, where $\Gamma= (\cov(\Psi(X; V)))^{-1}$. This is because for any
such pair $(s,t)$, we may form a junction tree with two nodes, one
containing~$s$ and one containing $t$, and apply
Corollary~\ref{CorSep}. Indeed, the matrix $\Gamma$ defined over
singleton vertices is agnostic to which triangulation we choose for
the graph.

In settings where there exists a junction tree representation of the
graph with only singleton separator sets, Corollary~\ref{CorTree} has
a number of useful implications for the consistency of methods that
have traditionally only been applied for edge recovery in Gaussian
graphical models: for
tree-structured discrete graphs, it suffices to estimate the support
of $(\cov(\Psi(X;V)))^{-1}$ from the data. We will review methods for
Gaussian graphical model selection and describe their analogs for
discrete tree graphs in Sections~\ref{SecGlobal}
and~\ref{SecNodewise}.

\subsection{Generalized covariances and neighborhood structure}

Theorem~\ref{ThmFull} also has a corollary, that is, relevant for
nodewise neighborhood selection approaches to graph
selection~\cite{MeiBuh06,RavEtal11}, which are applicable to graphs
with arbitrary topologies. Nodewise methods use the basic observation
that recovering the edge structure of $\graph$ is equivalent to
recovering the neighborhood set $N(s) = \{t \in\vertex\dvtx  (s,t) \in
\edge\}$ for each vertex $s \in\vertex$. For a given node $s \in
\vertex$ and positive integer $d$, consider the collection of subsets
\[
\mathcal{S}(s; d) \defn \bigl\{ U \subseteq\vertex\setminus\{s\}, |U| = d
\bigr\}.
\]
The following corollary provides an avenue for recovering $N(s)$ based
on the inverse of a certain generalized covariance matrix:

\begin{cor}[(Neighborhood selection)]\label{CorNeigh}
For any graph and node $s
\in\vertex$ with $\deg(s) \le d$, the inverse $\Gamma$ of the matrix
$\cov(\Psi (X; \{s\} \cup\operatorname{pow}(\mathcal{S}(s; d))))$ is
$s$-block graph-structured, that is, $\Gamma(\{s\}, B) = 0$ whenever
$\{s\} \neq B \subsetneq N(s)$. In particular, $\Gamma(\{s\}, \{t\}) =
0$ for all vertices $t \notin N(s)$.
\end{cor}

Note that $\operatorname{pow}(\mathcal{S}(s;d))$ is the set of subsets
of all candidate neighborhoods of~$s$ of size $d$. This result follows
from Theorem~\ref{ThmFull} (and the related Corollary~\ref{CorSep}) by
constructing a particular junction tree for the graph, in which $s$ is
separated from the rest of the graph by $N(s)$. Due to the well-known
relationship between the rows of an inverse covariance matrix and
linear regression coefficients~\cite{MeiBuh06},
Corollary~\ref{CorNeigh} motivates the following neighborhood-based
approach to graph selection: for a fixed vertex $s \in\vertex$, perform
a single \emph{linear regression} of $\Psi(X; \{s\})$ on the vector
$\Psi(X; \operatorname{pow}(\mathcal{S}(s; d)))$. Via elementary
algebra and an application of Corollary~\ref{CorNeigh}, the resulting
regression vector will expose the neighborhood $N(s)$ in an arbitrary
discrete graphical model; that is, the indicators $\Psi(X; \{t\})$
corresponding to $X_t$ will have a nonzero weight only if $t \in N(s)$.
We elaborate on this connection in Section~\ref{SecNodewise}.


\subsection{\texorpdfstring{Proof of Theorem~\protect\ref{ThmFull}}{Proof of Theorem 1}}
\label{SecProofThmFull}

We now turn to the proof of Theorem~\ref{ThmFull}, which is based on
certain fundamental correspondences arising from the theory of
exponential families~\cite{Barndorff78,Brown86,WaiJor08}. Recall that
our exponential family~\eqref{EqnGeneralMultinomial} has binary-valued
indicator functions~\eqref{EqnSuff} as its sufficient statistics. Let
$D$ denote the cardinality of this set and let $\mathbb{I}\dvtx
\mathcal{X}^p\rightarrow\{0,1\}^D$ denote the multivariate function
that maps each configuration $x \in\mathcal{X}^p$ to the vector
$\mathbb{I}(x)$ obtained by evaluating the $D$ indicator functions on
$x$. Using this notation, our exponential family may be written in the
compact form $q_\theta(x) = \exp\{\langle\theta, \mathbb{I}(x)\rangle-
\Phi(\theta)\}$, where
\begin{eqnarray*}
\bigl\langle\theta, \mathbb{I}(x) \bigr\rangle& =& \sum
_{C \in\mathcal{C}} \bigl\langle\theta_C, \mathbb{I}_C(x)
\bigr\rangle= \sum_{C \in\mathcal
{C}} \sum
_{J
\in\mathcal{X}_0^{|C|}} \theta_{C;J} \mathbb{I}_{C;J}(x_C).
\end{eqnarray*}
Since this exponential family is known to be minimal, we are
guaranteed~\cite{DarSpe83} that
\[
\nabla\Phi(\theta) = \Exs_\theta \bigl[\mathbb{I}(X) \bigr]
\quad\mbox{and}\quad\nabla^2 \Phi(\theta) = \cov_\theta
\bigl[\mathbb{I}(X) \bigr],
\]
where $\Exs_\theta$ and $\cov_\theta$ denote (resp.) the expectation
and covariance taken under the density
$q_\theta$~\cite{Brown86,WaiJor08}. The conjugate dual~\cite{Roc70} of
the cumulant function is given by
\[
\Phi^*(\mu) \defn\sup_{\theta\in\real^D} \bigl\{\langle\mu, \theta \rangle
- \Phi(\theta) \bigr\}.
\]
The function $\Phi^*$ is always convex and takes values in $\real
\cup\{+\infty\}$. From known results~\cite{WaiJor08}, the dual function
$\Phi^*$ is finite only for $\mu\in\real^D$ belonging to the marginal
polytope
%
\begin{equation}
\label{EqnMarginalPolytope} \mathcal{M} \defn \biggl\{\mu\in\real^p \biggm| \exists
 \mbox{ some density $q$ s.t. } \sum_{x} q(x)
\mathbb{I}(x) = \mu \biggr\}.
\end{equation}
%

The following lemma, proved in the supplementary material \cite{LohWai13Sup}, provides
a~connection between the covariance matrix and the Hessian of $\Phi^*$:
%
\begin{lem}
\label{LemEntropy} Consider a regular, minimal exponential family, and
define $\mu = \Exs_\theta[\mathbb{I}(X)]$ for any fixed
$\theta\in\Omega= \{\theta\dvtx  \Phi(\theta) < \infty\}$. Then
%
\begin{eqnarray}
\bigl(\cov_\theta \bigl[\mathbb{I}(X) \bigr] \bigr)^{-1} & =&
\nabla^2 \Phi^*(\mu).
\end{eqnarray}
\end{lem}

Note that the minimality and regularity of the family implies that
$\cov_\theta[\mathbb{I}(X)]$ is strictly positive definite, so the
matrix is invertible.\vadjust{\goodbreak}

For any $\mu\in\intr(\mathcal{M})$, let $\theta(\mu) \in\real^D$ denote
the unique natural parameter $\theta$ such that $\nabla\Phi(\theta) =
\mu$. It is known~\cite{WaiJor08} that the negative dual function
$-\Phi^*$ is linked to the Shannon entropy via the relation
%
\begin{eqnarray}
\label{EqnDualEntropy} - \Phi^*(\mu) & =& H \bigl(q_{\theta(\mu)}(x) \bigr) = -\sum
_{x \in
\mathcal{X}^p} q_{\theta(\mu)}(x) \log q_{\theta(\mu)}(x).
\end{eqnarray}
In general, expression~\eqref{EqnDualEntropy} does \emph{not} provide a
straightforward way to compute $\nabla^2 \Phi^*$, since the mapping
$\mu\mapsto\theta(\mu)$ may be extremely complicated. However, when the
exponential family is defined with respect to a triangulated graph,
$\Phi^*$~has an explicit closed-form representation in terms of the
mean parameters $\mu$. Consider a junction tree triangulation of the
graph, and let $(\widebar{\mathcal{C}}, \mathcal{S})$ be the collection
of maximal cliques and separator sets, respectively. By the junction
tree theorem \mbox{\cite{Lauritzen88,WaiJor08,KolFri09}}, we have the
factorization
%
\begin{equation}
\label{EqnJT} q(x_1, \ldots, x_p) = \frac{\prod_{C \in\widebar{\mathcal{C}}}
q_C(x_C)}{\prod_{S
\in
\mathcal{S}} q_S(x_S)},
\end{equation}
where $q_C$ and $q_S$ are the marginal distributions over maximal
clique $C$ and separator set $S$. Consequently, the entropy may be
decomposed into the sum
%
\begin{eqnarray}
\label{EqnPlainEntropy} H(q) & =& - \sum_{x \in\mathcal{X}^p} q(x) \log q(x)
= \sum_{C
\in\widebar{\mathcal{C}}} H_C(q_C) -
\sum_{S \in\mathcal{S}} H_S(q_S),
\end{eqnarray}
where we have introduced the clique- and separator-based entropies
\[
H_S(q_S) \defn- \sum_{x_S \in\mathcal{X}^{|S|}}
q_S(x_S) \log q_S(x_S)
\]
and
\[
H_C(q_C) \defn- \sum
_{x_C \in\mathcal{X}^{|C|}} q_C(x_C) \log
q_C(x_C).
\]

Given our choice of sufficient statistics~\eqref{EqnSuff}, we show
that $q_C$ and $q_S$ may be written explicitly as ``local'' functions
of mean parameters associated with $C$~and~$S$. For each subset $A
\subseteq\vertex$, let $\mu_A \in(\statenum-1)^{|A|}$ be the
associated collection of mean parameters, and let
\[
\mu_{\operatorname{pow}(A)} \defn\{ \mu_B \mid\varnothing\neq B \subseteq
A \}
\]
be the set of mean parameters associated with all nonempty subsets of
$A$. Note that $\mu_{\operatorname{pow}(A)}$ contains a total of
$\sum_{k=1}^{|A|} {|A| \choose k} (\statenum-1)^k = \statenum^{|A|} -
1$ parameters, corresponding to the number of degrees of freedom
involved in specifying a marginal distribution over the random vector
$x_A$. Moreover, $\mu_{\operatorname{pow}(A)}$ uniquely determines the
marginal distribution $q_A$:
%
\begin{lem}
\label{LemMapping} For any marginal distribution $q_A$ in the
$\statenum^{|A|}$-dimensional probability simplex, there is a unique
mean parameter vector $\mu_{\operatorname{pow}(A)}$ and matrix $M_A$
such that $q_A = M_A \cdot\mu_{\operatorname{pow}(A)}$.\vadjust{\goodbreak}
\end{lem}

For the proof, see the supplementary material \cite{LohWai13Sup}.


We now combine the dual representation~\eqref{EqnDualEntropy} with the
decomposition~\eqref{EqnPlainEntropy}, along with the matrices $\{M_C,
M_S\}$ from Lemma~\ref{LemMapping}, to conclude that
%
\begin{eqnarray}
\label{EqnEntropyExp} - \Phi^*(\mu) & =& \sum_{C \in\widebar{\mathcal{C}}}
H_C \bigl(M_C \bigl(\mu_{\operatorname{pow}(C)} \bigr) \bigr)
- \sum_{S \in
\mathcal{S}} H_S \bigl(M_S
\bigl(\mu_{\operatorname{pow}(S)} \bigr) \bigr).
\end{eqnarray}
%
Now consider two subsets $A, B \in\widetilde{\mathcal{C}}$ that are
not contained in
the same maximal clique. Suppose $A$ is contained within maximal
clique $C$. Differentiating expression~\eqref{EqnEntropyExp} with
respect to $\mu_A$ preserves only terms involving $q_C$ and $q_S$,
where $S$~is any separator set such that $A \subseteq S \subseteq
C$. Since $B \subsetneq C$, we clearly cannot have $B \subseteq
S$. Consequently, all cross-terms arising from the clique $C$ and its
associated separator sets vanish when we take a second derivative with
respect to $\mu_B$. Repeating this argument for any other maximal
clique $C'$ containing $A$ but not $B$, we have
$\frac{\partial^2 \Part^*}{\partial\mu_A\, \partial\mu_B}(\mu) = 0$.
This\vspace*{1pt} proves part (a).

Turning to part (b), note that if $A$ and $B$ are in the same maximal
clique, the expression obtained by taking second derivatives of the
entropy results in an algebraic expression with only finitely many
solutions in the parameters $\mu$ (consequently, also $\theta$). Hence,
assuming the $\theta$'s are drawn from a continuous distribution, the
corresponding values of the block $\Gamma(A,B)$ are a.s. nonzero.



\section{Consequences for graph structure estimation}
\label{SecSample}

Moving beyond the population level, we now state and prove several
results concerning the statistical consistency of different
methods---both known and some novel---for graph selection in discrete
graphical models, based on i.i.d. draws from a discrete graph. For
sparse Gaussian models, existing methods that exploit sparsity of the
inverse covariance matrix fall into two main categories: global graph
selection methods (e.g.,~\cite{AspBanGha08,FriEtal08,Rot08,RavEtal11})
and local (nodewise) neighborhood selection
methods~\cite{MeiBuh06,Zhao06}. We divide our discussion accordingly.


\subsection{Graphical Lasso for singleton separator graphs}
\label{SecGlobal}

We begin by describing how a combination of our population-level
results and some concentration inequalities may be leveraged to
analyze the statistical behavior of log-determinant methods for
discrete graphical models with singleton separator sets, and suggest
extensions of these methods when observations are systematically
corrupted by noise or missing data. Given a $p$-dimensional
random vector $(X_1, \ldots, X_p)$ with covariance $\Sigma^*$,
consider the estimator
%
\begin{equation}
\label{EqnLogDet} \widehat{\Theta}\in\arg\min_{\Theta\succeq0} \biggl\{
\trace(\widehat{\Sigma}\Theta) - \log\det(\Theta) + \lambda_n\sum
_{s \neq t} |\Theta_{st}| \biggr\},
\end{equation}
where $\widehat{\Sigma}$ is an estimator for $\Sigma^*$. For
multivariate Gaussian data, this program is an $\ell_1$-regularized
maximum likelihood estimate known as the \emph{graphical Lasso} and is
a well-studied method for recovering the edge structure in a Gaussian
graphical model~\cite{BanEtal08,FriEtal08,YuaLin07,Rot08}. Although the
program~\eqref{EqnLogDet} has no relation to the MLE in the case of a
discrete graphical model, it may still be useful for estimating
$\Theta^* \defn(\Sigma^*)^{-1}$. Indeed, as shown in Ravikumar et
al.~\cite{RavEtal11}, existing analyses of the
estimator~\eqref{EqnLogDet} require only tail conditions such as
sub-Gaussianity in order to guarantee that the sample minimizer is
close to the population minimizer. The analysis of this paper completes
the missing link by guaranteeing that the population-level inverse
covariance is in fact graph-structured. Consequently, we obtain the
interesting result that the program~\eqref{EqnLogDet}---even though it
is ostensibly derived from Gaussian considerations---is a consistent
method for recovering the structure of any binary graphical model with
singleton separator sets.

In order to state our conclusion precisely, we introduce additional
notation. Consider a general estimate $\widehat{\Sigma}$ of the
covariance matrix $\Sigma$ such that
%
\begin{eqnarray}
\label{EqnLogDetTreeCond} \mathbb{P} \biggl[\bigl\|\widehat{\Sigma}- \Sigma ^*\bigr\|_{\max}
\geq\varphi \bigl(\Sigma^* \bigr) \sqrt{\frac{\log p}{n}} \biggr] & \leq& c \exp
\bigl(- \psi(n, p) \bigr)
\end{eqnarray}
for functions $\varphi$ and $\psi$, where $\|\cdot\|_{\max}$ denotes
the elementwise $\ell_\infty$-norm. In the case of fully-observed
i.i.d. data with sub-Gaussian parameter $\sigma^2$, where
$\widehat{\Sigma}= \frac{1}{n} \sum_{i=1}^nx_i x_i^T - \bar{x}
\bar{x}{}^T$ is the usual sample covariance, this bound holds with
$\varphi(\Sigma^*) = \sigma^2$ and $\psi(n, p) = c' \log p$.

As in past analysis of the graphical Lasso~\cite{RavEtal11}, we require
a certain \emph{mutual incoherence} condition on the true covariance
matrix $\Sigma^*$ to control the correlation of nonedge variables with
edge variables in the graph. Let $\Gamma^*= \Sigma^* \otimes\Sigma^*$,
where $\otimes$ denotes the Kronecker product. Then $\Gamma^*$ is a
$p^2 \times p^2$ matrix indexed by vertex pairs. The incoherence
condition is given by
%
\begin{equation}
\label{EqnMatIncoherence} \max_{e \in S^c} \bigl\|\Gamma^*_{eS} \bigl(
\Gamma^*_{SS} \bigr)^{-1}\bigr\|_1 \le1-\alpha, \qquad
\alpha\in(0,1],
\end{equation}
where $S \defn\{(s,t)\dvtx  \Theta^*_{st} \neq0\}$ is the set of vertex
pairs corresponding to nonzero entries of the precision matrix
$\Theta^*$, equivalently, the edge set of the graph, by our theory on
tree-structured discrete graphs. For more intuition on the mutual
incoherence condition, see Ravikumar et al.~\cite{RavEtal11}.

With this notation, our global edge recovery algorithm proceeds as
follows:

%
\begin{alg}[(Graphical Lasso)]\label{AlgGraphLasso}
\begin{enumerate}
\item Form a suitable estimate $\widehat{\Sigma}$ of the true
    covariance matrix $\Sigma$.

\item Optimize the graphical Lasso program~\eqref{EqnLogDet} with
parameter $\lambda_n$, and denote the solution by $
\widehat{\Theta}$.

\item Threshold the entries of $\widehat{\Theta}$ at level $\tau_n$
to obtain an estimate of $\Theta^*$.
\end{enumerate}
\end{alg}

It remains to choose the parameters $(\lambda_n, \tau_n)$. In the
following corollary, we will establish statistical consistency of
$\widehat{\Theta}$ under the following settings:
%
\begin{equation}
\label{EqnLambdaTau} \lambda_n \ge\frac{c_1}{\alpha}\sqrt{
\frac{\log p}{n}}, \qquad\tau_n = c_2 \biggl\{
\frac{c_1}{\alpha} \sqrt{\frac{\log p}{n}} + \lambda_n \biggr\},
\end{equation}
where $\alpha$ is the incoherence parameter in
inequality~\eqref{EqnMatIncoherence} and $c_1, c_2$ are universal
positive constants. The following result applies to
Algorithm~\ref{AlgGraphLasso} when $\widehat{\Sigma}$ is the sample
covariance
matrix and $(\lambda_n, \tau_n)$ are chosen as in
equations~\eqref{EqnLambdaTau}:

\begin{cor}
\label{CorGlobIsing}
Consider an Ising model~\eqref{EqnIsing} defined by an undirected
graph with singleton separator sets and with degree at most
$d$, and suppose that the mutual incoherence
condition~\eqref{EqnMatIncoherence} holds. With $n\succsim
d^2 \log p$ samples, there are universal constants $(c,
c')$ such that with probability at least $1 - c \exp(-c' \log p)$,
Algorithm~\ref{AlgGraphLasso} recovers all edges $(s,t)$ with
$|\Theta^*_{st}| > \tau/2$.
\end{cor}

The proof is contained in the supplementary material \cite{LohWai13Sup}; it is a
relatively straightforward consequence of Corollary~\ref{CorSep} and
known concentration properties of $\widehat{\Sigma}$ as an estimate of
the population covariance matrix. Hence, if $|\Theta^*_{st}| > \tau/2$
for all edges $(s,t) \in E$, Corollary~\ref{CorGlobIsing} guarantees
that the log-determinant method plus thresholding recovers the full
graph exactly.

In the case of the standard sample covariance matrix, a variant of the
graphical Lasso has been implemented by Banerjee, El Ghaoui and
d'Aspre\-mont \cite{BanEtal08}. Our analysis establishes consistency of the
graphical Lasso for Ising models on single separator graphs using
$n\succsim d^2 \log p$ samples. This lower bound on the sample size is
unavoidable, as shown by information-theoretic
analysis~\cite{SanWai12}, and also appears in other past work on Ising
models~\cite{RavEtal10,JalEtal11,AnaEtal11}. Our analysis also has a
\emph{cautionary message}: the proof of Corollary~\ref{CorGlobIsing}
relies heavily on the population-level result in
Corollary~\ref{CorTree}, which ensures that $\Theta^*$ is
graph-structured when $G$ has only singleton separators. For a general
graph, we have no guarantees that $\Theta^*$ will be graph-structured
[e.g., see panel~(b) in Figure~\ref{FigGraphs}], so the graphical
Lasso~\eqref{EqnLogDet} is \emph{inconsistent in general}.

On the positive side, if we restrict ourselves to tree-structured
graphs, the estimator~\eqref{EqnLogDet} is attractive, since it relies
only on an estimate $\widehat{\Sigma}$ of the population covariance
$\Sigma^*$ that satisfies the deviation
condition~\eqref{EqnLogDetTreeCond}. In particular, even when the
samples $\{x_i\}_{i=1}^n$ are contaminated by noise or missing data, we
may form a good estimate $\widehat{\Sigma}$ of $\Sigma^*$. Furthermore,
the program~\eqref{EqnLogDet} is always convex regardless of whether
$\widehat{\Sigma}$ is positive semidefinite.

As a concrete example of how we may correct the
program~\eqref{EqnLogDet} to handle corrupted data, consider the case
when each entry of $x_i$ is missing independently with probability
$\rho$, and the corresponding observations $z_i$ are zero-filled for
missing entries. A natural estimator is
%
\begin{equation}
\label{EqnMissData} \widehat{\Sigma}= \Biggl( \frac{1}{n} \sum
_{i=1}^n z_i z_i^T
\Biggr) \div M - \frac{1}{(1-\rho)^2} \bar{z}\bar{z}{}^T,
\end{equation}
where $\div$ denotes elementwise division by the matrix $M$ with
diagonal entries $(1-\rho)$ and off-diagonal entries $(1-\rho)^2$,
correcting for the bias in both the mean and second moment terms. The
deviation condition~\eqref{EqnLogDetTreeCond} may be shown to hold
w.h.p., where $\varphi(\Sigma^*)$ scales with $(1-\rho)$ (cf. Loh and
Wainwright~\cite{LohWai11a}). Similarly, we may derive an appropriate
estimator $\widehat{\Sigma}$ for other forms of additive or
multiplicative corruption.

Generalizing to the case of $m$-ary discrete graphical models with $m
> 2$, we may easily modify the program~\eqref{EqnLogDet} by replacing
the elementwise $\ell_1$-penalty by the corresponding group
$\ell_1$-penalty, where the groups are the indicator variables for a
given vertex. Precise theoretical guarantees follow from results on the
group graphical Lasso~\cite{JacEtal09}.


\subsection{Consequences for nodewise regression in trees}
\label{SecNodewise}

Turning to local neighborhood selection methods, recall the
neighborhood-based method due to Meinshausen and
B\"{u}hlmann~\cite{MeiBuh06}. In a Gaussian graphical model, the
column
corresponding to node $s$ in the inverse covariance matrix $\Gamma=
\Sigma^{-1}$ is a scalar multiple of $\tilde{\beta}=
\Sigma_{\setminus s, \setminus s}^{-1} \Sigma_{\setminus s, s}$, the
limit of the linear regression vector for $X_s$ upon~$X_{\setminus s}$.
Based on $n$ i.i.d. samples from a $p$-dimensional multivariate
Gaussian distribution, the support of the graph may then be estimated
consistently under the usual Lasso scaling $n \succsim d \log p$, where
$d = |N(s)|$.

Motivated by our population-level results on the graph structure of
the inverse covariance matrix (Corollary~\ref{CorTree}), we now
propose a method for neighborhood selection in a tree-structured
graph. Although the method works for arbitrary $m$-ary trees, we state
explicit results only in the case of the binary Ising model to avoid
cluttering our presentation.

The method is based on the following steps. For each node $s \in V$, we
first perform $\ell_1$-regularized linear regression of $X_s$ against
$X_{\setminus s}$ by solving the modified Lasso program
%
\begin{equation}
\label{EqnNoisyLasso} \hat{\beta}\in\arg\min_{\|\beta\|_1 \le b_0
\sqrt{k}} \biggl\{
\frac{1}{2} \beta^T \widehat{\Gamma}\beta- \hat{ \gamma}
{}^T \beta+ \lambda_n \|\beta\|_1 \biggr\},
\end{equation}
where $b_0 > \|\tilde{\beta}\|_1$ is a constant, $(\widehat
{\Gamma}, \hat{\gamma})$ are suitable estimators for
$(\Sigma_{\setminus s, \setminus s}, \Sigma_{\setminus s, s})$, and
$\lambda_n$ is an appropriate parameter. We then combine the
neighborhood estimates over all nodes via an AND operation [edge
$(s,t)$ is present if both $s$ and $t$ are inferred to be neighbors of
each other] or an OR operation (at least one of $s$ or $t$ is inferred
to be a neighbor of the other).

Note that the program~\eqref{EqnNoisyLasso} differs from the standard
Lasso in the form of the \mbox{$\ell_1$-}constraint. Indeed, the normal
setting of the Lasso assumes a linear model where the predictor and
response variables are linked by independent sub-Gaussian noise, but
this is not the case for $X_s$ and $X_{\setminus s}$ in a discrete
graphical model. Furthermore, the generality of the
program~\eqref{EqnNoisyLasso} allows it to be easily modified to handle
corrupted variables via an appropriate choice of $(\widehat{\Gamma},
\hat{\gamma})$, as in Loh and Wainwright~\cite{LohWai11a}.\vadjust{\goodbreak}

The following algorithm summarizes our nodewise regression procedure
for recovering the neighborhood set $N(s)$ of a given node $s$:
%
\begin{alg}[(Nodewise method for trees)]\label{AlgNodewise}
\begin{enumerate}
\item Form a suitable pair of estimators $(\widehat{\Gamma},
    \hat{\gamma})$ for covariance submatrices
    ($\Sigma_{\setminus s, \setminus s}$, $\Sigma_{\setminus s, s}$).

\item Optimize the modified Lasso program~\eqref{EqnNoisyLasso} with
parameter $\lambda_n$, and denote the solution by $\hat{\beta}$.

\item Threshold the entries of $\hat{\beta}$ at level $\tau_n$, and
define the estimated neighborhood set $\widehat{N(s)}$ as the
support of the thresholded vector.
\end{enumerate}
\end{alg}

In the case of fully observed i.i.d. observations, we choose
$(\widehat{\Gamma}, \hat{\gamma})$ to be the recentered estimators
%
\begin{equation}
\label{EqnLassoPair} (\widehat{\Gamma}, \hat{\gamma}) = \biggl(\frac
{X_{\setminus s}^T X_{\setminus
s}}{n} -
\bar{x}_{\setminus s} \bar{x}_{\setminus s}^T,
\frac{X_{\setminus s}^T X_s}{n} - \bar{x}_s \bar{x}_{\setminus
s} \biggr)
\end{equation}
and assign the parameters $(\lambda_n, \tau_n)$ according to the
scaling
%
\begin{equation}
\label{EqnLambdaTau2} \lambda_n \succsim\varphi\|\tilde{\beta}
\|_2 \sqrt{\frac
{\log p}{n}}, \qquad\tau_n \asymp\varphi
\|\tilde{\beta}\|_2 \sqrt{\frac
{\log p}{n}},
\end{equation}
where $\tilde{\beta}\defn\Sigma_{\setminus s, \setminus s}^{-1}
\Sigma_{\setminus s, s}$ and $\varphi$ is some parameter such that
$\langle x_i, u\rangle$ is sub-Gaussian with parameter $\varphi^2
\|u\|_2^2$ for any $d$-sparse vector $u$, and $\varphi$ is independent
of $u$. The following result applies to Algorithm~\ref{AlgNodewise}
using the pairs $(\widehat{\Gamma}, \hat{\gamma})$ and $(\lambda_n,
\tau_n)$ defined as in equations~\eqref{EqnLassoPair}
and~\eqref{EqnLambdaTau2}, respectively.

\begin{prop}
\label{PropIsing} Suppose we have i.i.d. observations $\{x_i\}_{i=1}^n$
from an Ising model and that $n \succsim\varphi^2
\max\{\frac{1}{\lambda_{\min}(\Sigma_x)}, \llvert\!\llvert\!\llvert{\Sigma_x^{-1}}\rrvert\!\rrvert\!\rrvert_\infty^2 \} d^2 \log p$.
Then there are universal constants $(c, c', c'')$ such that with
probability greater\vspace*{1pt} than $1 - c \exp(-c' \log p)$, for any node $s \in
V$, Algorithm~\ref{AlgNodewise} recovers all neighbors $t \in N(s)$ for
which $|\tilde{\beta}_t| \geq c'' \varphi\|\tilde{\beta}\|_2
\sqrt{\frac{\log p}{n}}$.
\end{prop}

We prove this proposition in the supplementary material \cite{LohWai13Sup}, as a corollary of
a more general theorem on the $\ell_\infty$-consistency of the
program~\eqref{EqnNoisyLasso} for estimating~$\tilde{\beta}$,
allowing for corrupted observations. The theorem builds upon the
analysis of Loh and Wainwright~\cite{LohWai11a}, introducing techniques
for $\ell_\infty$-bounds and departing from the framework of a linear
model with independent sub-Gaussian noise.

\begin{rem*}
Regarding the sub-Gaussian parameter $\varphi$ appearing in
Proposition~\ref{PropIsing}, note that we may always take $\varphi=
\sqrt{d}$, since $|x_i^T u| \le\|u\|_1 \le\sqrt{d} \|u\|_2$ when $u$ is
$d$-sparse and $x_i$ is a binary vector. This leads to a sample
complexity requirement of $n \succsim d^3 \log p$. We suspect that a
tighter analysis, possibly combined with assumptions about the
correlation decay of the graph, would\vadjust{\goodbreak} reduce the sample complexity to
the scaling $n \succsim d^2 \log p$, as required by other methods with
fully observed data~\cite{JalEtal11,AnaEtal11,RavEtal10}. See the
simulations in Section~\ref{SecSims} for further discussion.

For corrupted observations, the strength and type of corruption enters
into the factors $(\varphi_1, \varphi_2)$ appearing in the deviation
bounds~(C.2a) and~(C.2b) below, and
Proposition~\ref{PropIsing} has natural extensions to the corrupted
case. We emphasize that although analogs of
Proposition~\ref{PropIsing} exist for other methods of graph selection
based on logistic regression and/or mutual information, the
theoretical analysis of those methods does not handle corrupted data,
whereas our results extend easily with the appropriate
scaling.\vspace*{-3pt}
\end{rem*}

In the case of $m$-ary tree-structured graphical models with $m > 2$,
we may perform multivariate regression with the multivariate group
Lasso~\cite{OboEtal11} for neighborhood selection, where groups are
defined (as in the log-determinant method) as sets of indicators for
each node. The general relationship between the best linear predictor
and the block structure of the inverse covariance matrix follows from
block matrix inversion, and
from a population-level perspective, it suffices to perform
multivariate linear regression of all indicators corresponding to a
given node against all indicators corresponding to other nodes in the
graph. The resulting vector of regression coefficients has nonzero
blocks corresponding to edges in the graph. We may also combine these
ideas with the group Lasso for multivariate regression~\cite{OboEtal11}
to reduce the complexity of the algorithm.\vspace*{-3pt}

\subsection{Consequences for nodewise regression in general graphs}
\label{SecNonTree}

Moving on from tree-structured graphical models, our method suggests a
graph recovery method based on nodewise linear regression for general
discrete graphs. Note that by Corollary~\ref{CorNeigh}, the inverse of
$\cov(\Psi(X; \operatorname{pow}(\mathcal{S}(s;d))))$ is $s$-block
graph-structured, where $d$ is such that $|N(s)| \le d$. It suffices to
perform a single multivariate regression of the indicators $\Psi(X;
\{s\})$ corresponding to node $s$ upon the other indicators in $\Psi(X;
V \cup \operatorname{pow}(\mathcal{S}(s;d)))$.

We again make precise statements for the binary Ising model ($m=2$). In
this case, the indicators $\Psi(X; \operatorname{pow}(U))$
corresponding to a subset of vertices $U$ of size $d'$ are all
$2^{d'}-1$ distinct products of variables $X_u$, for $u \in U$. Hence,
to recover the $d$ neighbors of node $s$, we use the following
algorithm. Note that knowledge of an upper bound $d$ is necessary for
applying the algorithm.

\begin{alg}[(Nodewise method for general graphs)]\label{AlgNonTreeNoDecay}
\begin{enumerate}
\item Use the modified Lasso program~\eqref{EqnNoisyLasso} with a
    suitable choice of $(\widehat{\Gamma}, \hat{\gamma})$ and
    regularization parameter $\lambda_n$ to perform a linear
    regression of $X_s$ upon all products of subsets of variables
    of $X_{\setminus s}$ of size at most $d$. Denote the solution
    by $\hat{\beta}$.

\item Threshold the entries of $\hat{\beta}$ at level $\tau_n$, and
define the estimated neighborhood set $\widehat{N(s)}$ as the
support of the thresholded vector.\vadjust{\goodbreak}
\end{enumerate}
\end{alg}

Our theory states that at the population level, nonzeros in the
regression vector correspond exactly to subsets of $N(s)$. Hence, the
statistical consistency result of Proposition~\ref{PropIsing} carries
over with minor modifications. Since Algorithm~\ref{AlgNonTreeNoDecay}
is essentially a version of Algorithm~\ref{AlgNonTree} with the first
two steps omitted, we refer the reader to the statement and proof of
Corollary~\ref{CorNonTree} below for precise mathematical statements.
Note here that since the regression vector has $\order(p^d)$
components, \mbox{$2^d-1$} of which are nonzero, the sample complexity of
Lasso regression in step~(1) of Algorithm~\ref{AlgNonTreeNoDecay} is
$\order(2^d \log(p^d)) = \order(2^d \log p)$.

For graphs exhibiting correlation decay~\cite{BreEtal08}, we may reduce
the computational complexity of the nodewise selection algorithm by
prescreening the nodes of $V \setminus s$ before performing a
Lasso-based linear regression. We define the nodewise correlation
according to
\[
r_C(s,t) \defn\sum_{x_s, x_t} \bigl|
\mathbb{P}(X_s = x_s, X_t =
x_t) - \mathbb{P}(X_s = x_s)
\mathbb{P}(X_t = x_t)\bigr|
\]
and say that the graph exhibits \emph{correlation decay} if there exist
constants $\zeta, \kappa> 0$ such that
%
\begin{equation}
\label{EqnCorrDecay} r_C(s,t) > \kappa\qquad\forall(s,t) \in E\quad
\mbox{and}\quad r_C(s,t) \le\exp \bigl(-\zeta r(s,t) \bigr)
\end{equation}
for all $(s,t) \in V \times V$, where $r(s,t)$ is the length of the
shortest path between $s$ and $t$. With this notation, we then have
the following algorithm for neighborhood recovery of a fixed node $s$
in a graph with correlation decay:

\begin{alg}[(Nodewise method with correlation decay)]\label{AlgNonTree}
\begin{enumerate}
\item Compute the empirical correlations
\[
\hat{r}_C(s,t) \defn\sum_{x_s, x_t} \bigl|
\widehat{\mathbb{P}}(X_s = x_s, X_t =
x_t) - \widehat{\mathbb{P}}(X_s = x_s)
\widehat{\mathbb{P}}(X_t = x_t)\bigr|
\]
between $s$ and all other nodes $t \in V$, where $\widehat{\mathbb
{P}}$ denotes
the empirical distribution.

\item Let $\mathcal{C}\defn\{t \in V\dvtx  \hat{r}_C(s,t) >
\kappa /2\}$ be the candidate set of nodes with sufficiently
high correlation. (Note that $\mathcal{C}$ is a function of
both $s$ and $\kappa$ and, by convention, $s
\notin\mathcal{C}$.)

\item Use the modified Lasso program~\eqref{EqnNoisyLasso} with
    parameter $\lambda_n$ to perform a linear regression of $X_s$
    against $\mathcal{C}_d \defn\Psi(X; V
    \cup\operatorname{pow}(\mathcal{C}(s;d))) \setminus\{X_s\}$,
    the set of all products of subsets of variables $\{X_c\dvtx  c
    \in\mathcal{C}\}$ of size at most $d$, together with singleton
    variables. Denote the solution by $\hat{\beta}$.

\item Threshold the entries of $\hat{\beta}$ at level $\tau_n$, and
define the estimated neighborhood set $\widehat{N(s)}$ as the
support of the thresholded vector.
\end{enumerate}
\end{alg}

Note that Algorithm~\ref{AlgNonTreeNoDecay} is a version of
Algorithm~\ref{AlgNonTree} with $\mathcal{C}= V \setminus s$,
indicating the absence of a prescreening step. Hence, the
statistical
consistency result below applies easily to
Algorithm~\ref{AlgNonTreeNoDecay} for graphs with no correlation decay.\vadjust{\goodbreak}

For fully observed i.i.d. observations, we choose $(\widehat{\Gamma},
\hat{\gamma})$ according to
%
\begin{equation}
\label{EqnGamParam} (\widehat{\Gamma}, \hat{\gamma}) = \biggl(\frac
{X_\mathcal{C}
^T X_\mathcal{C}}{n}
- \bar{x}_\mathcal{C} \bar{x}_\mathcal{C}^T,
\frac
{X_\mathcal{C}^T X_{s}}{n} - \bar{x}_{s} \bar{x}_\mathcal{C}
\biggr)
\end{equation}
and parameters $(\lambda_n, \tau_n)$ as follows: for a candidate set
$\mathcal{C}$, let $x_{\mathcal{C}, i} \in\{0, 1\}^{|\mathcal{C}_d|}$
denote the augmented vector corresponding to the observation $x_i$, and
define $\Sigma_\mathcal{C}\defn\operatorname{Cov}(x_{\mathcal{C}, i},
x_{\mathcal{C}, i})$. Let
$\tilde{\beta}\defn\Sigma_\mathcal{C}^{-1}
\operatorname{Cov}(x_{\mathcal{C}, i}, x_{s,i})$. Then set
%
\begin{equation}
\label{EqnLambdaTau3} \lambda_n \succsim\varphi\|\tilde{\beta}
\|_2 \sqrt{\frac
{\log
|\mathcal{C}_d|}{n}}, \qquad\tau_n \asymp\varphi
\|\tilde{\beta}\|_2 \sqrt{\frac{\log|\mathcal{C}_d|}{n}},
\end{equation}
where $\varphi$ is some function such that $\langle x_{\mathcal{C}, i},
u\rangle$ is sub-Gaussian with parameter $\varphi^2 \|u\|_2^2$ for any
$(2^d - 1)$-sparse vector $u$, and $\varphi$ does not depend on $u$. We
have the following consistency result, the analog of
Proposition~\ref{PropIsing} for the augmented set of vectors. It
applies to Algorithm~\ref{AlgNonTree} with the pairs $(\widehat
{\Gamma}, \hat{\gamma})$ and $(\lambda_n, \tau_n)$ chosen as in
equations~\eqref{EqnGamParam} and~\eqref{EqnLambdaTau3}.

\begin{cor}
\label{CorNonTree} Consider i.i.d. observations $\{x_i\}_{i=1}^n$
generated from an Ising model satisfying the correlation decay
condition~\eqref{EqnCorrDecay}, and suppose
%
\begin{equation}
\label{EqnHam} n\succsim \biggl( \kappa^2 + \varphi^2
\max \biggl\{\frac{1}{\lambda_{\min}(\Sigma_\mathcal{C})}, \bigl\llvert\!\bigl\llvert\!\bigl\llvert
\Sigma_\mathcal{C}^{-1} \bigr\rrvert\!\bigr\rrvert\!\bigr\rrvert
_\infty^2 \biggr\} 2^{2d} \biggr) \log|
\mathcal{C}_d|.
\end{equation}
Then there are universal constants $(c, c', c'')$ such that with
probability at least $1-c \exp(-c' \log p)$, and for any $s \in
V$:
\begin{longlist}[(ii)]
\item[(i)] The set $\mathcal{C}$ from step (2) of
    Algorithm~\ref{AlgNonTree} satisfies $|\mathcal{C}| \le
    d^{(\log(4/\kappa))/\zeta}$.

\item[(ii)] Algorithm~\ref{AlgNonTree} recovers all neighbors $t \in
N(s)$ such that
\[
|\tilde{\beta}_t| \geq c'' \varphi
\|\tilde{\beta}\|_2 \sqrt{\frac{\log|\mathcal{C}_d|}{n}}.
\]
\end{longlist}
\end{cor}

The proof of Corollary~\ref{CorNonTree} is contained in
the supplementary material \cite{LohWai13Sup}. Due to the exponential factor $2^d$
appearing in the lower bound~\eqref{EqnHam} on the sample size, this
method is suitable only for bounded-degree graphs. However, for
reasonable sizes of $d$, the dimension of the linear regression problem
decreases from $\order(p^d)$ to $|\mathcal{C}_d| =
\order(|\mathcal{C}|^d) = \order(d^{(d \log(4/\kappa))/\zeta} )$, which
has a significant impact on the runtime of the algorithm. We explore
two classes of bounded-degree graphs with correlation decay in the
simulations of Section~\ref{SecSims}, where we generate
Erd\"{o}s--Renyi graphs with edge probability $c/p$ and square grid
graphs in order to test the behavior of our recovery algorithm on
nontrees. When $m > 2$, corresponding to nonbinary states, we may
combine these ideas with the overlapping group Lasso~\cite{JacEtal09}
to obtain similar algorithms for nodewise recovery of nontree graphs.
However, the details are more complicated, and we do not include them
here. Note that our method for nodewise recovery in nontree graphical
models is again easily adapted to handle noisy and missing data, which
is a clear advantage over other existing methods.


\subsection{Simulations}
\label{SecSims}

In this section we report the results of various simulations we
performed to illustrate the sharpness of our theoretical claims. In all
cases, we generated data from binary Ising models. We first applied the
nodewise linear regression method (Algorithm \ref{AlgNodewise} for trees; Algorithm \ref{AlgNonTreeNoDecay}
in the general case) to the method of $\ell_1$-regularized logistic
regression, analyzed in past work for Ising model selection by
Ravikumar, Wainwright and Lafferty
\cite{RavEtal10}. Their main result was to establish
that, under certain incoherence conditions of the Fisher information
matrix, performing \mbox{$\ell_1$-}regularized logistic regression with a
sample size $n\succsim d^3 \log p$ is guaranteed to select the correct
graph w.h.p. Thus, for any bounded-degree graph, the sample size $n$
need grow only logarithmically in the number of nodes $p$. Under this
scaling, our theory also guarantees that nodewise \emph{linear
regression} with $\ell_1$-regularization will succeed in recovering the
true graph w.h.p.

%
\begin{figure}
\begin{tabular}{@{}c@{\quad}c@{}}

\includegraphics{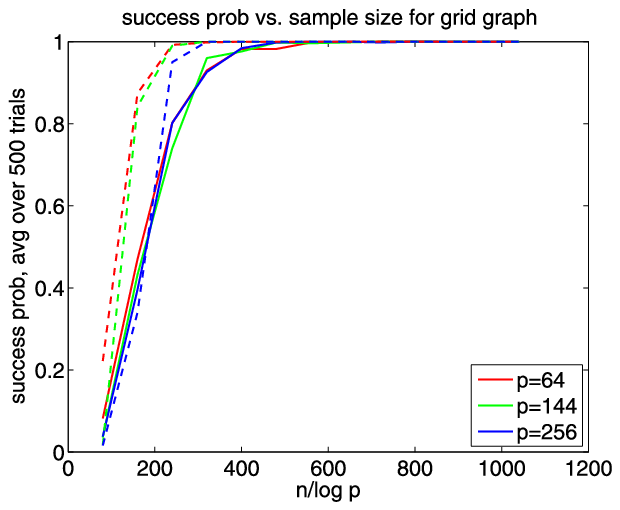}
 & \includegraphics{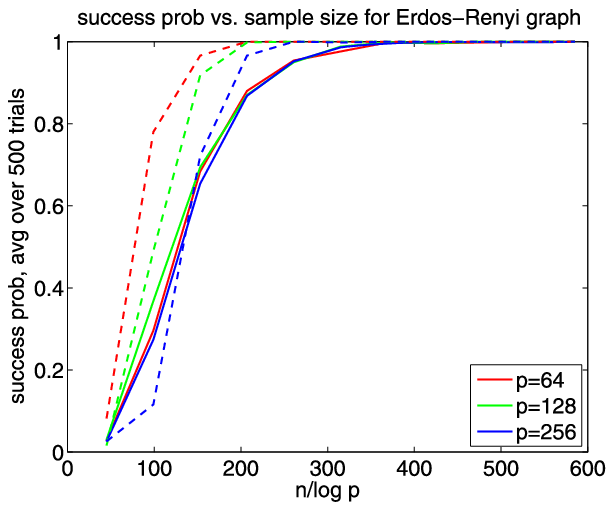}\\
\footnotesize{(a) Grid graph} & \footnotesize{(b) Erd\"{o}s--Renyi graph}\\[6pt]
\multicolumn{2}{@{}c@{}}{
\includegraphics{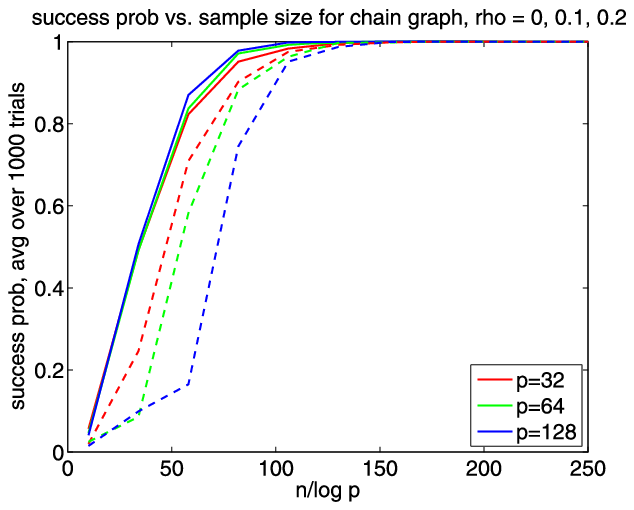}
}\\
\multicolumn{2}{@{}c@{}}{\footnotesize{(c) Chain graph}}
\end{tabular}
\caption{Comparison between $\ell_1$-regularized logistic vs. linear
regression methods for graph recovery. Each panel plots of the
probability of correct graph recovery vs. the rescaled sample size
$n/\log p$; solid curves correspond to linear regression
(method in this paper), whereas dotted curves correspond to logistic
regression~\cite{RavEtal10}. Curves are based on average
performance over 500 trials. \textup{(a)}~Simulation results for
two-dimensional grids with $d= 4$ neighbors, and number of
nodes $p$ varying over $\{64, 144, 256 \}$. Consistent with
theory, when plotted vs. the rescaled sample size $n/\log
p$, all three curves (red, blue, green) are well aligned with
one another. Both linear and logistic regression transition from
failure to success at a similar point. \textup{(b)} Analogous results for an
Erd\"{o}s--Renyi graph with edge probability $3/p$. \textup{(c)} Analogous
results for a chain-structured graph with maximum degree $d=2$.}\label{FigSimsOne}
\end{figure}

In Figure~\ref{FigSimsOne} we present the results of simulations with
two goals: (i) to test the $n\approx\log p$ scaling of the required
sample size; and (ii) to compare $\ell_1$-regularized nodewise linear
regression (Algorithms \ref{AlgNonTreeNoDecay} and \ref{AlgNonTree}) to $\ell_1$-regularized nodewise
logistic regression~\cite{RavEtal10}. We ran simulations for the two
methods on both tree-structured and nontree graphs with data generated
from a binary Ising model, with node weights $\theta_s = 0.1$ and edge
weights $\theta_{st} = 0.3$. To save on computation, we employed the
neighborhood screening method described in Section~\ref{SecNonTree} to
prune the candidate neighborhood set before performing linear
regression. We selected a~candidate neighborhood set of size
$\lfloor2.5 d \rfloor$ with highest empirical correlations, then
performed a single regression against all singleton nodes and products
of subsets of the candidate neighborhood set of size at most $d$, via
the modified Lasso program~\eqref{EqnNoisyLasso}. The size of the
candidate neighborhood set was tuned through repeated runs of the
algorithm. For both methods, the optimal choice of regularization
parameter $\lambda_n$ scales as $\sqrt{\frac{\log p}{n}}$, and we used
the same value of $\lambda_n$ in comparing logistic to linear
regression. In each panel we plot the probability of successful graph
recovery versus the rescaled sample size $\frac{n}{\log p}$, with
curves of different colors corresponding to graphs (from the same
family) of different sizes. Solid lines correspond to linear
regression, whereas dotted lines correspond to logistic regression;
panels (a), (b) and (c) correspond to grid graphs, Erd\"{o}s--Renyi
random graphs and chain graphs, respectively. For all these graphs, the
three solid/dotted curves for different problem sizes are well aligned,
showing that the method undergoes a transition from failure to success
as a function of the ratio $\frac{n}{\log p}$. In addition, both linear
and logistic regression are comparable in terms of statistical
efficiency (the number of samples $n$ required for correct graph
selection to be achieved).

%
\begin{figure}
\begin{tabular}{@{}cc@{}}

\includegraphics{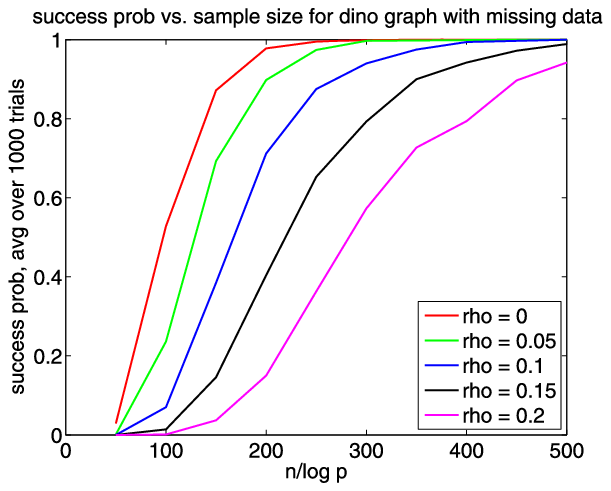}
 & \includegraphics{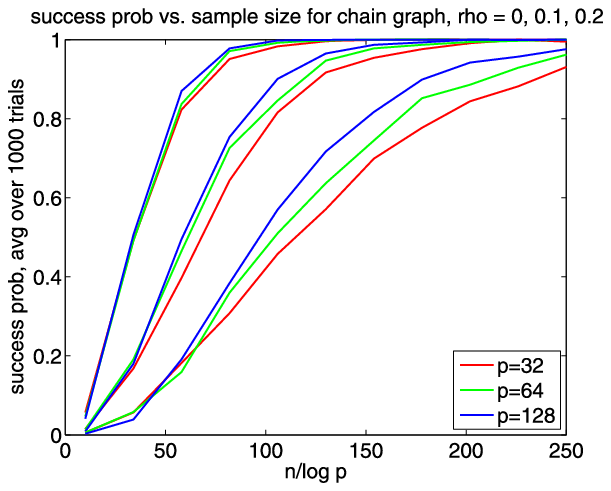}\\
\footnotesize{(a) Dino graph with missing data} & \footnotesize{(b) Chain graph with missing data}\\[6pt]
\multicolumn{2}{@{}c@{}}{
\includegraphics{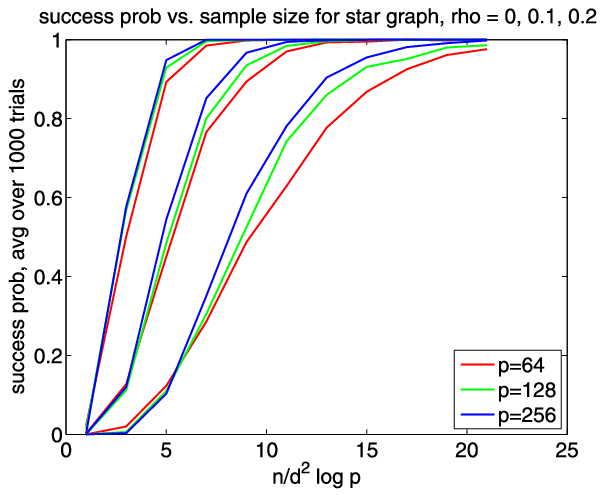}
}\\
\multicolumn{2}{@{}c@{}}{\footnotesize{(c) Star graph with missing data, $d \approx\log p$}}
\end{tabular}
\caption{Simulation results for global and nodewise recovery methods
on binary Ising models, allowing for missing data in the
observations. Each point represents an average over 1000
trials. Panel~\textup{(a)} shows simulation results for the graphical Lasso
method applied to the dinosaur graph with the fraction $\rho$ of
missing data varying in $\{0, 0.05, 0.1, 0.15, 0.2\}$. Panel \textup{(b)}
shows simulation results for nodewise regression applied to chain
graphs for varying $p$ and $\rho$. Panel \textup{(c)} shows simulation
results for nodewise regression applied to star graphs with maximal
node degree $d= \log p$ and varying $\rho$.}\label{FigSimsTwo}
\end{figure}

The main advantage of nodewise linear regression and the graphical
Lasso over nodewise logistic regression is that they are
straightforward to correct for corrupted or missing data.
Figure~\ref{FigSimsTwo} shows the results of simulations designed to
test the behavior of these corrected estimators in the presence of
missing data. Panel~(a) shows the results of applying the graphical
Lasso method, as described in Section~\ref{SecGlobal}, to the dino
graph of Figure~\ref{FigGraphs}(e). We again generated data from an
Ising model with node weights $0.1$ and edge weights $0.3$. The curves
show the probability of success in recovering the 15 edges of the
graph, as a function of the rescaled sample size $\frac{n}{\log p}$ for
$p = 13$. In addition, we performed simulations for different levels of
missing data, specified by the parameter $\rho\in \{0, 0.05, 0.1, 0.15,
0.2\}$, using the corrected estimator~\eqref{EqnMissData}. Note that
all five runs display a transition from success probability 0 to
success probability 1 in roughly the same range, as predicted by our
theory. Indeed, since the dinosaur graph has only singleton \mbox{separators},
Corollary~\ref{CorTree} ensures that the inverse covariance matrix is
exactly graph-structured, so our global recovery method is consistent
at the population level. Further note that the curves shift right as
the fraction $\rho$ of missing data increases, since the recovery
problem becomes incrementally harder.

Panels (b) and (c) of Figure~\ref{FigSimsTwo} show the results of the
nodewise regression method of Section~\ref{SecNodewise} applied to
chain and star graphs, with increasing numbers of nodes $p \in\{32, 64,
128\}$ and $p \in\{64, 128, 256\}$, respectively. For the chain graphs
in panel (b), we set node weights of the Ising model equal to $0.1$ and
edge weights equal to $0.3$. For the varying-degree star graph in panel
(c), we set node weights equal to $0.1$ and edge weights equal to
$\frac{1.2}{d}$, where the degree $d$ of the central hub grows with the
size of the graph as $\lfloor\log p \rfloor$. Again, we show curves for
different levels of missing data, $\rho\in\{0, 0.1, 0.2\}$. The
modified Lasso program~\eqref{EqnNoisyLasso} was optimized using a form
of composite gradient descent due to Agarwal, Negahban and Wainwright
\cite{AgaEtal11}, guaranteed to converge to a small neighborhood of the optimum even when
the problem is nonconvex~\cite{LohWai11a}. In both the chain and star
graphs, the three curves corresponding to different problem sizes $p$
at each value of the missing data parameter $\rho$ stack up when
plotted against the rescaled sample size. Note that the curves for the
star graph stack up nicely\vspace*{-1pt} with the scaling $\frac{n}{d^2 \log p}$,
rather than the worst-case scaling $n \asymp d^3 \log p$, corroborating
the remark following Proposition~\ref{PropIsing}. Since $d = 2$ is
fixed for the chain graph, we use the rescaled sample size
$\frac{n}{\log p}$ in our plots, as in the plots in
Figure~\ref{FigSimsOne}. Once again, these simulations corroborate our
theoretical predictions: the corrected linear regression estimator
remains consistent even in the presence of missing data, although the
sample size required for consistency grows as the fraction of missing
data $\rho$ increases.


\section{Discussion}
\label{SecDiscussion}

The correspondence between the inverse covariance matrix and graph
structure of a Gauss--Markov random field is a classical fact with
numerous consequences for estimation of Gaussian graphical models. It
has been an open question as to whether similar properties extend to a
broader class of graphical models. In this paper, we have provided a
partial affirmative answer to this question and developed theoretical
results extending such relationships to discrete undirected graphical
models.

As shown by our results, the inverse of the ordinary covariance matrix
is graph-structured for special subclasses of graphs with singleton
separator sets. More generally, we have considered inverses of
\emph{generalized covariance matrices}, formed by introducing
indicator functions for larger subsets of variables. When these
subsets are chosen to reflect the structure of an underlying junction
tree, the edge structure is reflected in the inverse covariance
matrix. Our population-level results have a number of statistical
consequences for graphical model selection. We have shown that our
results may be used to establish consistency (or inconsistency) of
standard methods for discrete graph selection, and have proposed new
methods for neighborhood recovery which, unlike existing methods, may
be applied even when observations are systematically corrupted by
mechanisms such as additive noise and missing data. Furthermore, our
methods are attractive in their simplicity, in that they only involve
simple optimization problems.

\section*{Acknowledgments}
Thanks
to the Associate Editor and anonymous reviewers for helpful feedback.

\begin{supplement}
\stitle{Supplementary material for ``Structure estimation for discrete graphical models: Generalized
covariance matrices and~their~inverses''}
\slink[doi,text=10.1214/13-AOS1162SUPP]{10.1214/13-AOS1162SUPP} 
\sdatatype{.pdf}
\sfilename{aos1162\_supp.pdf}
\sdescription{Due to space constraints, we have relegated technical details of the
remaining proofs to the supplement~\cite{LohWai13Sup}.}
\end{supplement}

%

\printaddresses

\end{document}